\documentclass[letterpaper, 10 pt, conference]{ieeeconf2}
\usepackage{multicol}
\usepackage[bookmarks=true]{hyperref}

\usepackage{xcolor}
\usepackage{algorithm}
\usepackage{algpseudocode}
\usepackage{caption}
\usepackage{amsmath}
\usepackage{graphicx}
\usepackage{tikz}
\usepackage{xspace}
\usepackage{listings}
\usepackage{tabularx}
\usepackage{booktabs}

\usepackage{pifont}
\definecolor{greentick}{RGB}{89, 161, 79}
\definecolor{redtick}{RGB}{225, 87, 89}
\newcommand{\cmark}{\textcolor{greentick}{\ding{51}}}%
\newcommand{\xmark}{\textcolor{redtick}{\ding{55}}}%

\newcommand{\name}{PEERNet\xspace}

\IEEEoverridecommandlockouts                              
\title{\LARGE \bf \name: An End-to-End Profiling Tool for 
Real-Time \\ Networked Robotic Systems}

\begin{document}

\author{\authorblockN{Aditya Narayanan, Pranav Kasibhatla, Minkyu Choi, Po-han Li, Ruihan Zhao, and Sandeep Chinchali
\thanks{The University of Texas at Austin.}
\thanks{Corresponding author: {\tt\small adityan@utexas.edu}}}}

\maketitle

\begin{abstract}
Networked robotic systems balance compute, power, and latency constraints in applications such as self-driving vehicles, drone swarms, and teleoperated surgery.
A core problem in this domain is deciding when to offload a computationally expensive task to the cloud, a remote server, at the cost of communication latency. 
Task offloading algorithms often rely on precise knowledge of system-specific performance metrics, such as sensor data rates, network bandwidth, and machine learning model latency. 
While these metrics can be modeled during system design, uncertainties in connection quality, server load, and hardware conditions introduce real-time performance variations, hindering overall performance.
    We introduce \name, an end-to-end and real-time profiling tool for cloud robotics.
\name enables performance monitoring on heterogeneous hardware through targeted yet adaptive profiling of system components such as sensors, networks, deep-learning pipelines, and devices.
We showcase \name's capabilities through networked robotics tasks, such as image-based teleoperation of a Franka Emika Panda arm and querying vision language models using an Nvidia Jetson Orin.
\name reveals non-intuitive behavior in robotic systems, such as asymmetric network transmission and bimodal language model output.
Our evaluation underscores the effectiveness and importance of benchmarking in networked robotics, demonstrating \name's adaptability. Our code is open-source and available at \href{https://github.com/UTAustin-SwarmLab/PEERNet}{{\tt\small github.com/UTAustin-SwarmLab/PEERNet}}.
\end{abstract}

\section{Introduction}
Networked robotics \cite{kehoe2015survey, mao2017survey} is a proven framework for addressing compute, power, and latency constraints in robotic systems, with applications in self-driving vehicles \cite{kumar2012cloud}, drone swarms \cite{Asaamoning2021DroneSA}, and teleoperated surgery \cite{berkelman2009compact}. 
A key challenge in networked robotics is to decide when to offload computationally expensive tasks to a remote server at the cost of network latency.
Previous works have studied optimal strategies for data sharing and computation offloading in various networked robotics settings \cite{chinchali2019network, ghosh2022dynamic, li2024online, lin2020survey, safeNetworkedRobot_Narasimhan}. 
However, these algorithms face an issue in deployment: they lack real-time data on system performance. Without real-time monitoring of dynamic real-world systems, networked robotics is difficult to optimize.

For example, consider an autonomous vehicle that offloads perception and localization tasks to a remote server, the so-called cloud, over a wireless network connection. Some practical system design questions include:
\begin{enumerate}
    \item How can we select a deep learning model that ensures decision accuracy while minimizing latency?
    \item How is the overall latency affected by the type of connectivity, \textit{e.g.}, Wi-Fi, LTE vs. 5G?
    \item What models of LiDAR and cameras provide data rates fast enough to match the network bandwidth?
\end{enumerate}
Answering such questions requires a granular understanding of the performance of individual system components, from sensor data rates to Machine Learning (ML) inference latency to network transmission delay.
This practical engineering use case shows that real-time system-specific performance profiling is essential for the efficient deployment of modern networked robotics.

Conventional methods for profiling networked robotic systems fall into two categories: isolated benchmarking and end-to-end system benchmarking. Existing end-to-end system benchmarks \cite{mayoralvilches2024robotperf, caldas2019leaf} assess components within fully assembled systems, but frequently overlook networking, especially in asymmetric networks where upload and download latencies differ. These benchmarks generally fail to present a holistic software platform for modular profiling.

\begin{figure}
    \centering
    \vspace{0.75em}
    \includegraphics[width = \linewidth]{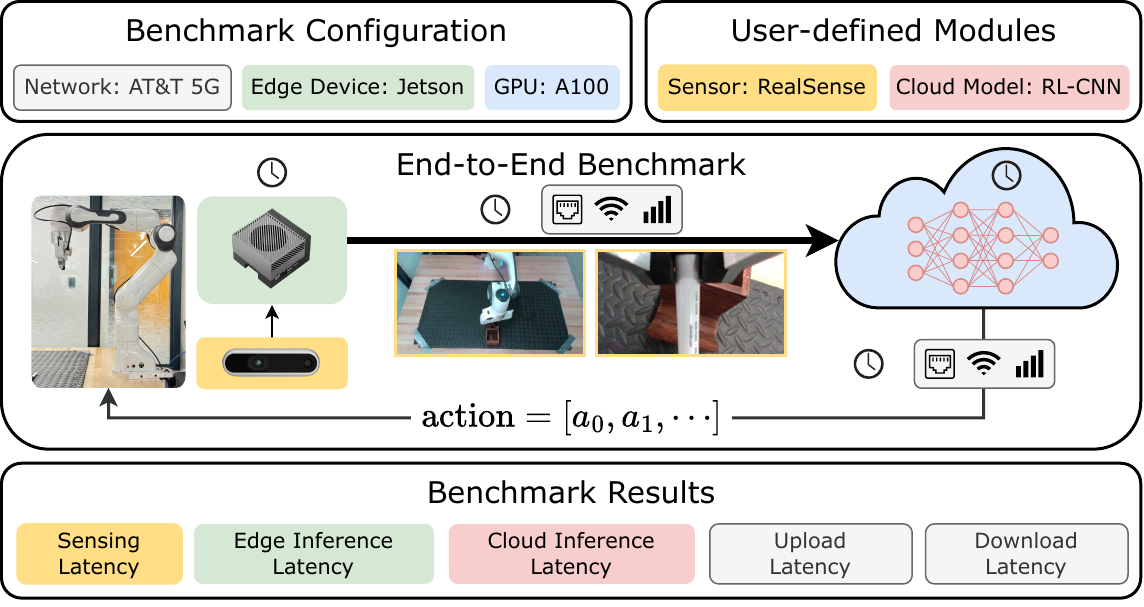}
    \caption{\small \textbf{\name is an end-to-end profiling tool for networked robotics.} 
    Our modular tool allows users to inject custom external modules and specify custom configurations of sensors, networks, deep-learning pipelines and devices. 
    Detailed profiles of latency, complete with asymmetric network timing, are generated.
    }
    \label{fig:system-overview}
    \vspace{-2em}
\end{figure}

For these reasons, we introduce the \textbf{P}rofiler for \textbf{E}nd-to-\textbf{E}nd \textbf{R}eal-time \textbf{Net}worked robotic systems (\name), a novel end-to-end profiling tool designed for real-time networked robotic systems. By modeling arbitrary networked robotic systems as combinations of sensors, networks, and computation, our method is capable of detailed profiling of arbitrary user-defined systems. This is the first tool that integrates with industry-standard products such as Nvidia single board computers and Robot Operating System (ROS), allowing users to systematically benchmark all aspects of a cloud or edge robotics system.
As shown in Figure \ref{fig:system-overview}, our system emphasizes end-to-end profiling of the entire flow of data from sensing to upload, remote inference, and download, allowing us to profile at a level of detail not possible with isolated benchmarks.

In summary, our contributions are:
\begin{enumerate}
    \item \textbf{End-to-End Networked Robotics Benchmarking:} We present \name, a Python framework for end-to-end benchmarking of networked robotics systems, profiling data flow and system performance.
    
    \item \textbf{Enhanced Modularity and Accessibility:} \name boasts a high degree of modularity and accessibility, facilitated by highly modular implementation and a comprehensive Command-Line Interface (CLI) which allows users to define custom networked robotics setups to profile, even allowing users to pass external programs into the benchmarking framework.

    \item \textbf{One-way Delay Estimation:} Catered towards networked robotics, \name times one-way network delay using Network Time Protocol (NTP) synchronization. Compared to round-trip time measurements, \name provides specificity in the data collected, especially those of networked robotics systems operating in asymmetric networks. For example, our experiments show that the latency variations between upload and download are notable because edge devices upload considerably more data than they download.
   
    \item \textbf{Deployment Examples on Real Systems:}
    We explore three distinct robotics scenarios encompassing offloaded inference and teleoperated robotics. Leveraging the capabilities of \name, we conduct comprehensive benchmarking to assess performance across various implementations of networked robotics. Our findings underscore the end-to-end benchmarking capabilities of our framework, emphasizing its prowess in profiling, extensibility, and modularity.
\end{enumerate}

\section{Related Works}
This section discusses progress towards benchmarking systems for networked robotics.
\subsection{Standardized Tasks and Datasets}
One approach to benchmarking in modern robotics is to develop standardized tasks and datasets to provide meaningful comparisons between robotic systems. RTRBench \cite{RTRBench} provides standardized ``kernels" for common robotic perception and planning tasks. In the context of networked robotics, IoTBench \cite{IoTBench} outlines the requirements and difficulties of producing a comprehensive benchmark for networked systems. Live robotics profiling remains an open challenge, as existing approaches ignore frameworks for data collection.

\subsection{Isolated Component Benchmarking}
Other approaches focus on profiling individual components in robotic systems, \textit{i.e.}, sensors or ML models. Baidu's DeepBench \cite{baidu_deepbench} provides a flexible framework for profiling operations in ML inference, and MLPerf \cite{reddi2020mlperf} builds one of the largest databases of crowd-sourced ML inference benchmarks. Aside from the more prevalent benchmarking tools and datasets for ML tasks, recent works also provide datasets and data collection methods for hardware components, such as LiDAR \cite{schalling2019benchmarking}. While these isolated benchmarks can be comprehensive due to crowd-sourcing, they do not truly reflect the holistic performance of robotic systems, as they do not reflect data flow relationships seen in real hardware.

\subsection{Network Delay Estimation}
Estimating network delay is a crucial component in profiling networked robotic systems. Network delay estimation is well studied, both in the round-trip case \cite{acharya1998study} and the one-way delay case \cite{de2008one, choi2005one}. In this work, we integrate one-way delay estimation using clock synchronization \cite{smotlacha2003one} into a networked robotics profiling framework.

\subsection{System-level Benchmarking Tools}
In recent years, many holistic system-level benchmarking tools have been developed. RobotPerf \cite{mayoralvilches2024robotperf} is a holistic benchmarking framework on top of ROS2 \cite{ROS2}, and PyRobot \cite{pyrobot} provides a high-level interface to facilitate research and benchmarking with ROS systems. LEAF \cite{caldas2019leaf}, a software package tailored to benchmark federated learning, closely approaches networked robotic benchmarking by considering multi-robot systems and the effects of networking. Despite a system-level holistic view of robotic systems, these tools fail to provide a general solution for networked robotics benchmarking by ignoring network latency and being constrained to particular hardware or software stacks.

FogRos \cite{chen2021fogros, ichnowski2023fogros2} builds a lightweight framework for easy deployment of ROS applications to cloud architectures. Similarly, \name allows interfacing with standard software stacks such as ROS and ROS2, but aims to add profiling capabilities to cloud robotic systems by considering diverse network types and arbitrary components.

\subsection{Teleoperation Latency Studies}
In robotic teleoperation, prior works have quantified network latency and its effects on teleoperation task performance \cite{lum_teleoperation, cundar_latency}. However, these methods focus on particular cases of networked robotic systems rather than building tools for general and modular networked robotics. 

In conclusion, existing methods and tools for robotic system benchmarking do not fully address the problem of networked robotics profiling. Although standardized tasks compare individual systems, they lack insight into the system's implementations. Isolated component benchmarks are easy to use and comprehensive but fail to reflect the combined performance of completed systems. Network profiling, including one-way delay estimation, is well documented, but prior work has not integrated it into the benchmarking of networked robotics.

\section{A Motivating Example}
\label{problem-statement}
To motivate the need for \name, we present a practical networked robotics example in which end-to-end and real-time benchmarking is essential.
The example includes common functional components in networked robotic systems, \textit{e.g.}, sensors, networks, and devices with computing capabilities, which can all be profiled by \name.

Consider an automated factory where quality control is accomplished using computer vision models trained to identify defects in parts on a production line. The pipeline for such a system is as follows: A camera (sensor) samples images from videos from an assembly line. Due to computing constraints, these images are either processed locally on the robot arm's (devices) low-level controller or offloaded over the Internet (network) to cloud servers such as Amazon Web Services machines (devices). In the cloud, images are classified using computationally expensive vision models, and labels are transmitted back to the factory. Finally, the robot arm can discard defective parts. 

To deploy offloading algorithms minimizing end-to-end latency of such systems, the following must be known:
\begin{itemize}
    \item Data rates of sensors: the frequency at which images of the assembly line can be sampled.
    \item Local inference latency: the time taken to classify an image as defective or not on the local device.
    \item Network upload latency: the time taken to transmit images to the cloud.
    \item Cloud inference latency: the time taken to classify an image as defective or not on the cloud.
    \item Network download latency: the time taken to transmit labels back to the robot arms.
    \item Robot arm latency: the time taken for the robot arms to discard a sample.
\end{itemize}
Complete knowledge of these metrics in real-time enables optimization of this motivating example and more general networked robotic systems. \name presents a modular and easy-to-use software stack to collect these metrics.

\section{Implementation of \name}
In Section \ref{problem-statement}, we provide an example of a networked robotic system that requires real-time profiling to optimize. We now present the implementation of the main modules of \name, closely following that motivating example.

\subsection{Functional Modules} \label{functional-modules}
\name consists of four major functional modules that closely mirror the components of the networked robotic example presented in Section \ref{problem-statement}. Our software stack provides sample implementations of each module but more importantly, a framework for users to customize.

\begin{enumerate}
    \item The \textbf{Sensors} module abstracts input data streams to edge devices, defining sensors as any hardware or software from which a program can sample data. It includes traditional sensors such as cameras and Inertial Measurement Units (IMUs) and non-traditional input streams such as datasets or human input.
    \item The \textbf{Networks} module abstracts all communication between devices. Our base software stack contains three network implementations frequently used in networked robotic systems: a Transmission Control Protocol (TCP) implementation built through ZeroMQ (a lightweight messaging library), a User Datagram Protocol (UDP) implementation built through ZeroMQ, and a wrapper for networking in ROS \cite{quigley2009ros}. The network module also contains a set of utilities for transmitting and receiving metadata that enables one-way network profiling with time-synced devices.
    \item The \textbf{Inference} module abstracts computation on local and cloud devices. For example, implementations of the inference module encapsulate object detection, image classification, and Large Language Model (LLM) queries. However, the framework we provide is general and encapsulates general computing tasks.
    \item The \textbf{Logging} module is responsible for organizing and maintaining the data collected during end-to-end benchmarks. The core component of the logging module is the logger, a tree-like data structure with serializable nodes for tracking iterative and nested components of a benchmark. Nodes in a logger represent events during the life cycle of a benchmark. We implement a data-typing system for nodes, with nodes typed based on the metric they collect, such as latency or throughput. The logger provides an interface for users to define custom node types to represent custom metrics, such as power consumption or temperature measurements. As detailed in Section \ref{network-timing}, the serializable and recursive structure of this data structure enables concise profiling of one-way transmission time among time-synced devices.
\end{enumerate}

\subsection{Implementation Details}
\begin{figure}
    \centering
    \includegraphics[width = \linewidth]{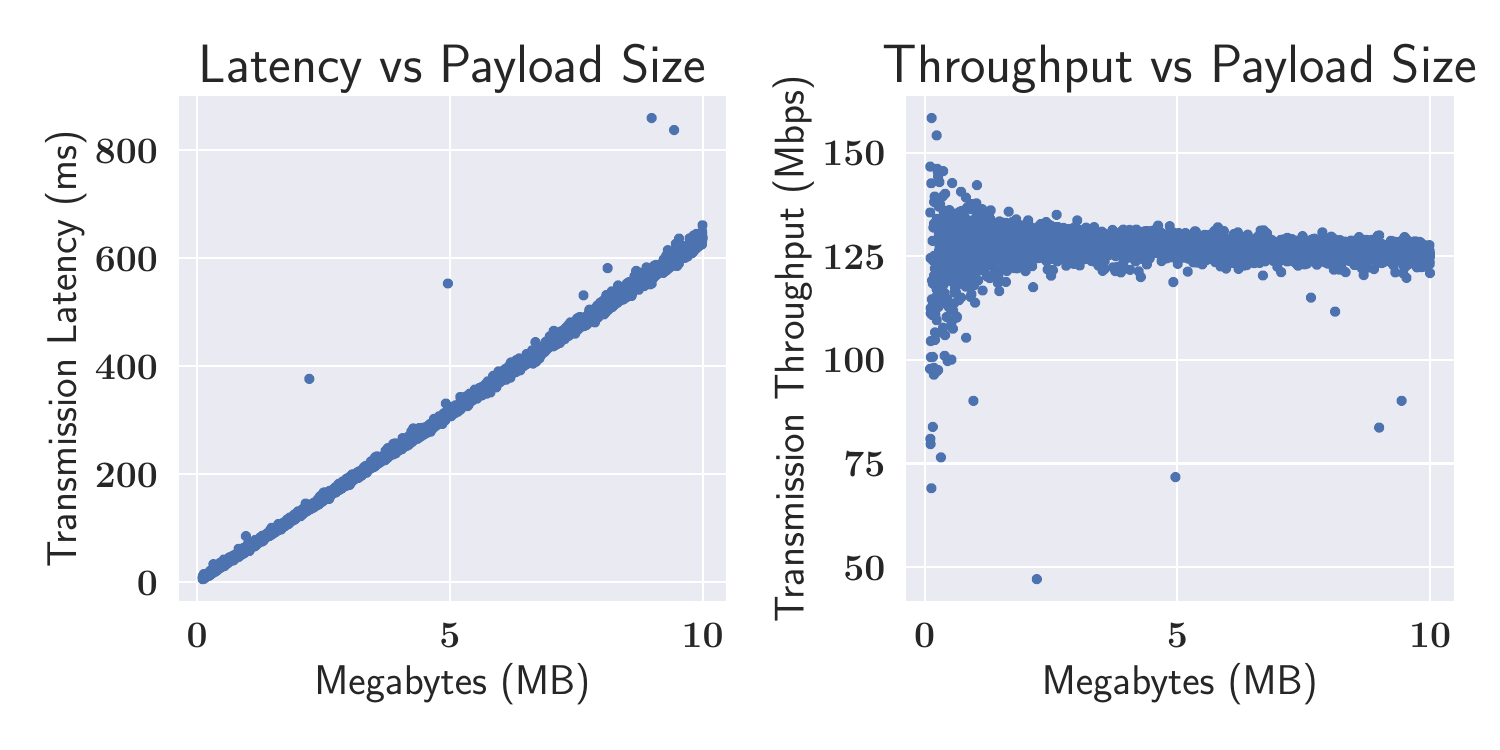}
    \caption{\small \textbf{\name's one-way delay measurements incur minimal error, of $2.01$ms.} Latency estimates exhibit a constant offset error, measured here to be $-2.01$ms, and throughput estimates show decreasing variance for higher payload sizes, in the scale of images.}
    \label{fig:latency-throughput}
    \vspace{-1em}
\end{figure}

We now present how \name composes the four functional modules described in Section \ref{functional-modules} to make an end-to-end benchmarking tool. We cover the network timing scheme used to profile one-way delay in Figure \ref{fig:latency-throughput} and the implementation of a CLI to benchmark offloaded computation.

\subsubsection{One-Way Network Profiling} \label{network-timing}
In many networked robotic systems, devices exhibit asymmetric data transfer; edge devices upload much more data than they download, and servers see more incoming traffic \cite{balakrishnan1997effects}. In our smart factory example, images (on the scale of megabytes each) are uploaded to the cloud, but only labels (on the scale of kilobytes) are downloaded. 
Thus, \name implements a one-way delay profiling scheme. Our implementation involves syncing device clocks on the sender and receiver using an NTP implementation such as Chrony, as seen in \cite{smotlacha2003one}. With synced device clocks, we exploit the serializable structure of the nodes in our logger to augment all messages sent across networks with logging metadata. Devices on the receiving end can deserialize attached metadata and interact with logger objects. This allows \name to start the timing on a sending device before a message is sent and to end the timing on a receiving device after a message is received.

Measuring one-way delay using synchronized clocks incurs some error due to imperfect clock synchronization. However, we quantify the total error in one-way delay estimates and study its propagation to estimates of other network-related metrics such as throughput. Figure \ref{fig:latency-throughput} shows the results of an introspective study, where we observe the latency and throughput of various payload sizes. As the ideal relationship between latency and payload size is linear with the intercept $0$, we calculate the intercept of a line of best fit to estimate the total one-way delay estimation error, which is $-2.01$ ms. The negative intercept indicates that the sending device's clock is faster than the receiver's. The ratio between throughput and payload size is ideally constant. We show that for sufficiently large payloads such as images, it holds. For small payloads, the throughput estimates are affected by small offsets in the latency estimates.

\begin{algorithm}[t]
    \caption{\name's Offloaded Inference Scheme}
    \label{alg: offloaded-inference}
    \begin{algorithmic}[1]
        \State \textbf{Definitions:}
        \State $\mathcal{S}$: A sensor
        \State $\mathcal{D}$: A device that can sample data from $\mathcal{S}$
        \State $\mathcal{M}$: An ML model
        \State $\mathcal{C}$: A cloud device that can perform inference using $\mathcal{M}$
        \State $\mathcal{N}$: A bi-directional network between $\mathcal{D}$ and $\mathcal{C}$
        \bigskip
        \State \textbf{Offloaded Inference Scheme:}
        \While{not done}
        \State $\mathcal{D}$ samples data $x$ from $\mathcal{S}$
        \State $\mathcal{D}$ uploads $x$ to $\mathcal{C}$ over $\mathcal{N}$
        \State $\mathcal{C}$ performs inference, obtaining $y = \mathcal{M}(x)$
        \State $\mathcal{C}$ transmits $y$ to $\mathcal{D}$ over $\mathcal{N}$
        \EndWhile
    \end{algorithmic}
\end{algorithm}

\subsubsection{CLI for Offloaded Inference}

Many networked robotics systems, including our smart factory example, fall into the category of offloaded computation or offloaded inference \cite{chinchali2019network, ghosh2022dynamic, li2024online, lin2020survey, safeNetworkedRobot_Narasimhan}. In these systems, edge devices selectively offload computation to cloud servers to conserve computing resources and minimize latency. \name provides an implementation that combines its functional modules into an offloaded edge robotic system (Algorithm \ref{alg: offloaded-inference}) and exposes a CLI, enabling users to define custom sensors, networks, and computing programs. The CLI is highly modular, allowing for custom selection of sensors, networks, and models, even allowing users to pass custom implementations of sensors and ML models from a single terminal command.

The pipeline for our offloaded inference implementation (Algorithm \ref{alg: offloaded-inference}) is as follows. First, users specify on the command line from an edge device $\mathcal{D}$, a cloud device $\mathcal{C}$, a sensor $\mathcal{S}$, a network type $\mathcal{N}$, and a cloud model $\mathcal{M}$. The edge device samples a datapoint $x$ from the sensor, and uploads $x$ to $\mathcal{C}$ over $\mathcal{N}$. Cloud device $\mathcal{C}$ performs inference with model $\mathcal{M}$, and returns the output $y$ to edge device $\mathcal{D}$ over network $\mathcal{N}$. During the process, \name's logging module is used to profile every operation indicated in algorithm \ref{alg: offloaded-inference} such as sampling from a sensor and one-way network transmission. Appendix \ref{app:CLI} contains a full list of metrics collected and examples of CLI usage.

In summary, \name is implemented around four core functional modules: sensors, networks, inference, and logging. It provides utilities for profiling one-way delay across networks using NTP for clock syncing and implements a CLI tailored to benchmark offloaded inference systems.

\section{Experimental Validation}
We validate \name by implementing and profiling three networked robotic systems. We demonstrate that \name is modular, robust, and capable of precisely profiling various systems. All of our experiments are conducted on physical hardware and live wireless networks. Our first experiment, offloaded image classification, demonstrates our profiler's generalizability to common networked robotics systems. Our second experiment demonstrates \name's profiling of complex behavior in LLM inference. Finally, our third experiment, a teleoperation example on a Franka Emika Panda robot arm, demonstrates \name's ability to profile arbitrary systems in a modular fashion.

\begin{figure}[t!]
    \centering
    \vspace{-1em}
    \includegraphics[width = \linewidth]{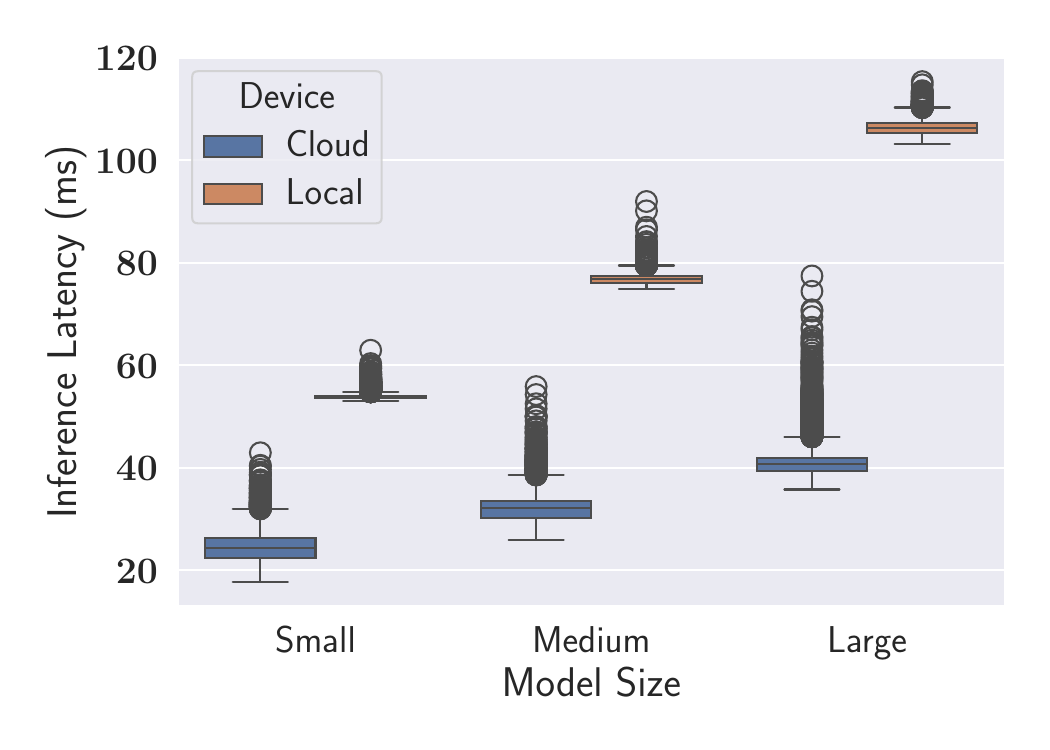}
    \caption{\small \textbf{\name precisely quantifies inference costs at the edge and in the cloud.} For the EfficientNetV2 family of models, local computation on an edge device is roughly 2.5 times as slow as offloaded computation to a cloud server, but cloud servers display a high variance in inference latency.}
    \label{fig:efficientnet-paired}
    \vspace{-1.5em}
\end{figure}
\begin{figure*}[t!]
    \centering
    \includegraphics[width = 0.90\linewidth]{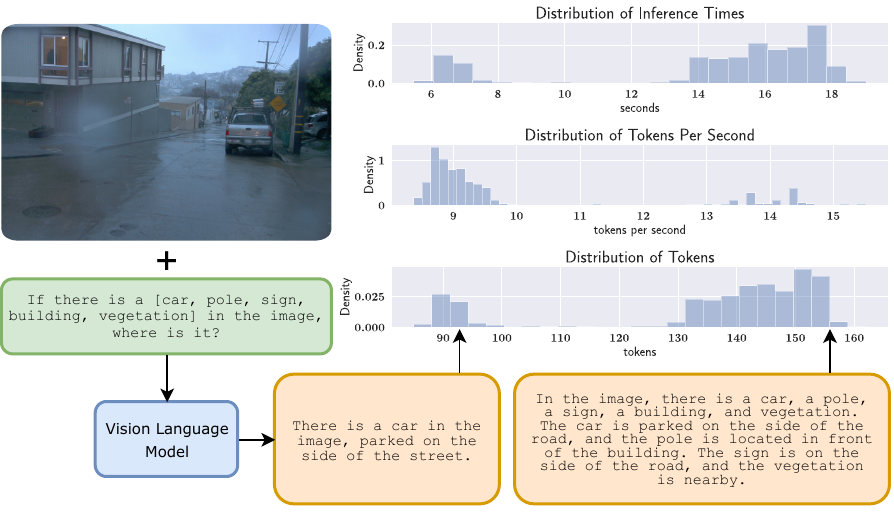}
    \caption{\small \textbf{Profiling with \name reveals non-intuitive behavior of a vision language model.} Responses to a single prompt vary in length, with two centers in the distribution of output length. Consequently, the inference latency is bimodal.}
    \label{fig:llava}
    \vspace{-2em}
\end{figure*}

\subsection{Offloaded Image Classification with EfficientNetV2}
Consider a self-driving car performing image classification in real-time on a video stream. For a resource-constrained edge device, offloading image classification to cloud servers is a viable solution. Suppose further that our self-driving car has access to a bank of image classification models, each with varying model complexities. To develop methods to select which image classifier from the bank to run, data on the inference latency of each model running locally and on cloud servers is essential. 

We implement such an offloaded image classification system with the following parameters. The edge device is an Nvidia Jetson AGX Orin 64Gb, a popular single-board computer for deep learning on the edge. The bank of models is the EfficientNetV2 family of image classifiers \cite{tan2021efficientnetv2}, trained on the ImageNet dataset \cite{deng2009imagenet}, which includes three model variants: small, medium, and large. A remote server with an Nvidia A5000 GPU plays the role of a cloud server. The remote server is configured as an NTP server, and the Nvidia Jetson is configured as an NTP client. 
As this is a case of offloaded computation, the experiment is fully specified through \name's CLI for offloaded inference. We specify a network type, provide domain names and IP addresses for each device, and specify a model to be run, either locally or in the cloud. Appendix \ref{app:terminal} contains the precise terminal commands issued to reproduce our offloaded image classification experiment.

Figure \ref{fig:efficientnet-paired} shows the performance of each variant of EfficientNetV2 when run locally and on a cloud server. \name's profiling enables us to precisely quantify the latency of each variant of EfficientNetV2 locally and on the cloud server. We see that each variant running locally is roughly $2.5$ times slower than the same model running in the cloud. A more subtle observation from the profiling is the high variance in inference latency in the cloud compared to the low variance in inference latency locally. This observation is consistent with the fact that cloud servers have many users, while edge devices rarely run many parallel processes. Thus, the variance in inference latency in the cloud is large compared to the variance in latency on the edge.
In summary, this experiment demonstrates the powerful abstraction of offloaded computation as a framework to benchmark in and validates \name's CLI as a modular and easy-to-use method to obtain precise profiling data on offloaded computation systems.

\subsection{LLM Inference at the Edge}
Recently, the robotics community has begun to adopt LLMs in robotic pipelines. These models provide reasoning capabilities, performing tasks such as planning \cite{sharan2024plan, vemprala2023chatgpt}. Furthermore, multi-modal LLMs such as LLaVA \cite{liu2024visual} are particularly suited to robotics applications due to their ability to reason about images and video. Due to the size of modern LLMs, many applications adopt a cloud inference model, where LLMs are served over wireless networks. However, with advancing edge hardware and powerful GPUs at the edge, the feasibility of LLM inference at the edge is an open question.

In this experiment, we use \name to profile the performance of LLaVA \cite{liu2024visual}, a multi-modal LLM capable of image reasoning on edge inference. We implement and benchmark the edge-inference system using \name's utilities, and demonstrate the ability to obtain complex information about model performance.
In our experimental setup, the edge device is an Nvidia Jetson AGX Orin 64Gb. The multi-modal LLM we use is LLaVA-v1.5-7b \cite{liu2023improved}. We focus our profiling on understanding the output length of LLM responses and the relationship between output length and inference latency.

Figure \ref{fig:llava} presents the collected data on the performance of LLM inference at the edge, along with a test prompt and image, from the Waymo Open Dataset \cite{Sun_2020_CVPR}. Benchmarking with \name reveals that for a particular image and prompt, LLaVA-v1.5's output length is strongly bimodal. Furthermore, we note that our edge device can produce between $8$ and $15$ tokens per second, corresponding to between $5$ and $20$ seconds per inference with a particular prompt. 

\subsection{Teleoperation of a Robotic Arm}

Teleoperation, in which a physical robot is controlled remotely by another machine or human, is a prevalent use case of networked robotics, with implementations in self-driving \cite{tener2022driving, zhang2020toward}, augmented reality \cite{walker2019robot}, and telesurgery \cite{sedaghat2021rt, barba2022remote}. We design, implement, and benchmark a teleoperation system for a Franka Emika Panda robot arm to demonstrate how \name can be used to profile complex teleoperation systems.

\begin{figure}[t!]
    \centering
    \includegraphics[width = \linewidth]{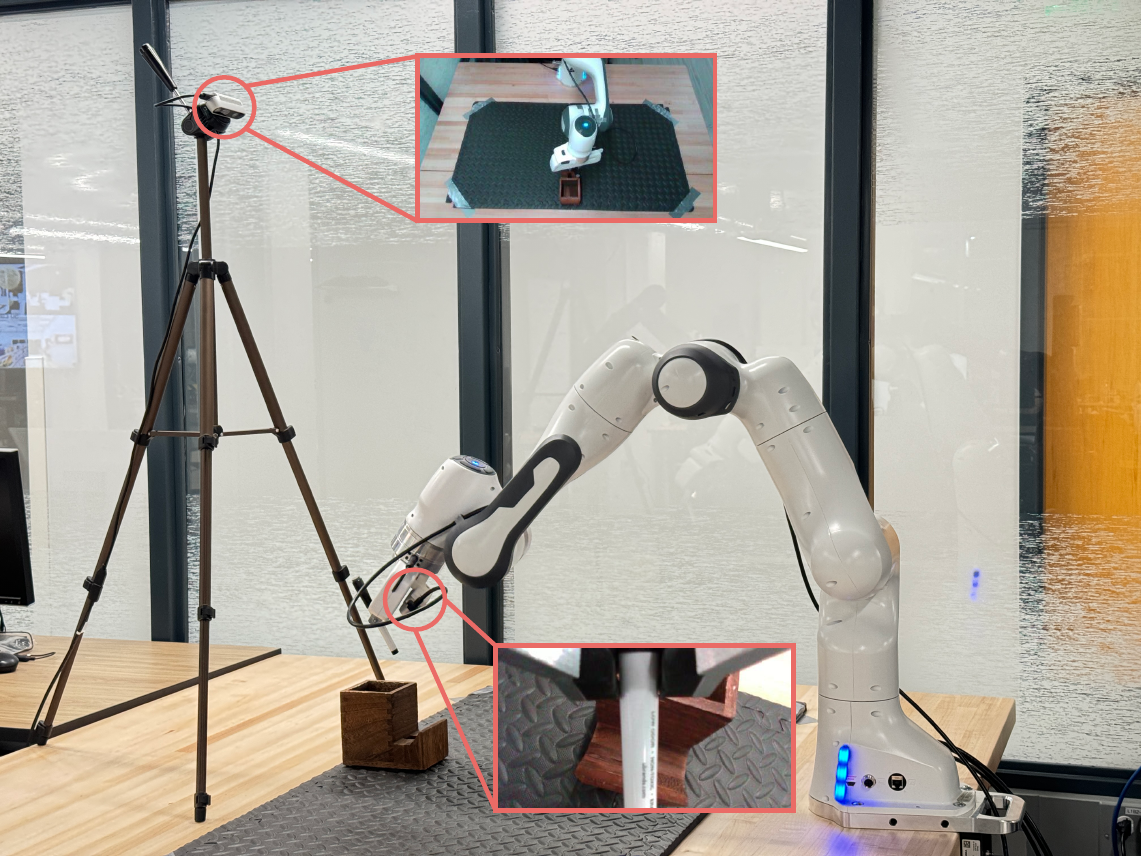}
    \caption{\small \textbf{Physical setup for Teleoperation.} The objective of teleoperation is for the robot arm to manipulate the pen into the wooden cup, using visual information from two cameras, one on the arm, and another on a tripod.}
    \label{fig: robot-arm}
    \vspace{-1em}
\end{figure}

We implement and profile a teleoperation pipeline of vision-based robot manipulation. In this task, the manipulator must start from its initial configuration $50$cm above the table and manipulate a pen into a square cup (7.5cm x 7.5cm), using image observations from two camera angles. We outfit a Franka Emika Panda robotic arm with two RealSense cameras, one mounted on the end effector and one mounted on a tripod (Figure \ref{fig: robot-arm}). We employ a CNN-based reinforcement learning policy to ensure consistency in task performance \cite{zhao2023learning}. 

We consider several different configurations to accomplish our teleoperation task, using different GPUs, networks, and image compression. We profile each configuration with \name to compare end-to-end performance and demonstrate how \name builds an understanding of sources of latency and their effects on end-to-end latency.
All teleoperation configurations follow a similar pipeline. An edge device (Nvidia Jetson Orin 64GB) samples images from the two RealSense cameras. Based on the configuration, the edge device may downsample images to lower the resolution and size, which we refer to as resizing. The edge device uploads images to a chosen cloud device over a chosen network type, where inference is performed with the CNN-based controller. Actions are downloaded back to the edge, where the robot arm executes them. This process is repeated until the episode terminates after a fixed number of steps. For each configuration, we profile $30$ episodes, each with $30$ steps until termination. We discard the first episode and the first step from each episode to account for GPU warm-up.

\par We consider three types of local and cloud devices:
\begin{enumerate}
    \item \textbf{Jetson}: For a \textit{local} configuration, the Jetson edge device performs inference locally using the GPU onboard.
    \item \textbf{RTX3090}: A desktop machine with an Nvidia RTX3090 GPU.
    \item \textbf{A100}: A remote server with an Nvidia A100 GPU.
\end{enumerate}

We consider three types of networks:
\begin{enumerate}
    \item \textbf{Local Area Network (LAN)}: Devices are connected via Ethernet to a network switch.
    \item \textbf{Campus Area Network (CAN)}: Devices are connected via WiFi to a campus-wide network.
    \item \textbf{AT\&T 5G}: Devices are connected to the internet via an AT\&T 5G mobile hotspot, and connected to a CAN through a Virtual Private Network (VPN).
\end{enumerate}

\begin{figure}[t!]
    \centering
    \includegraphics[width=\linewidth]{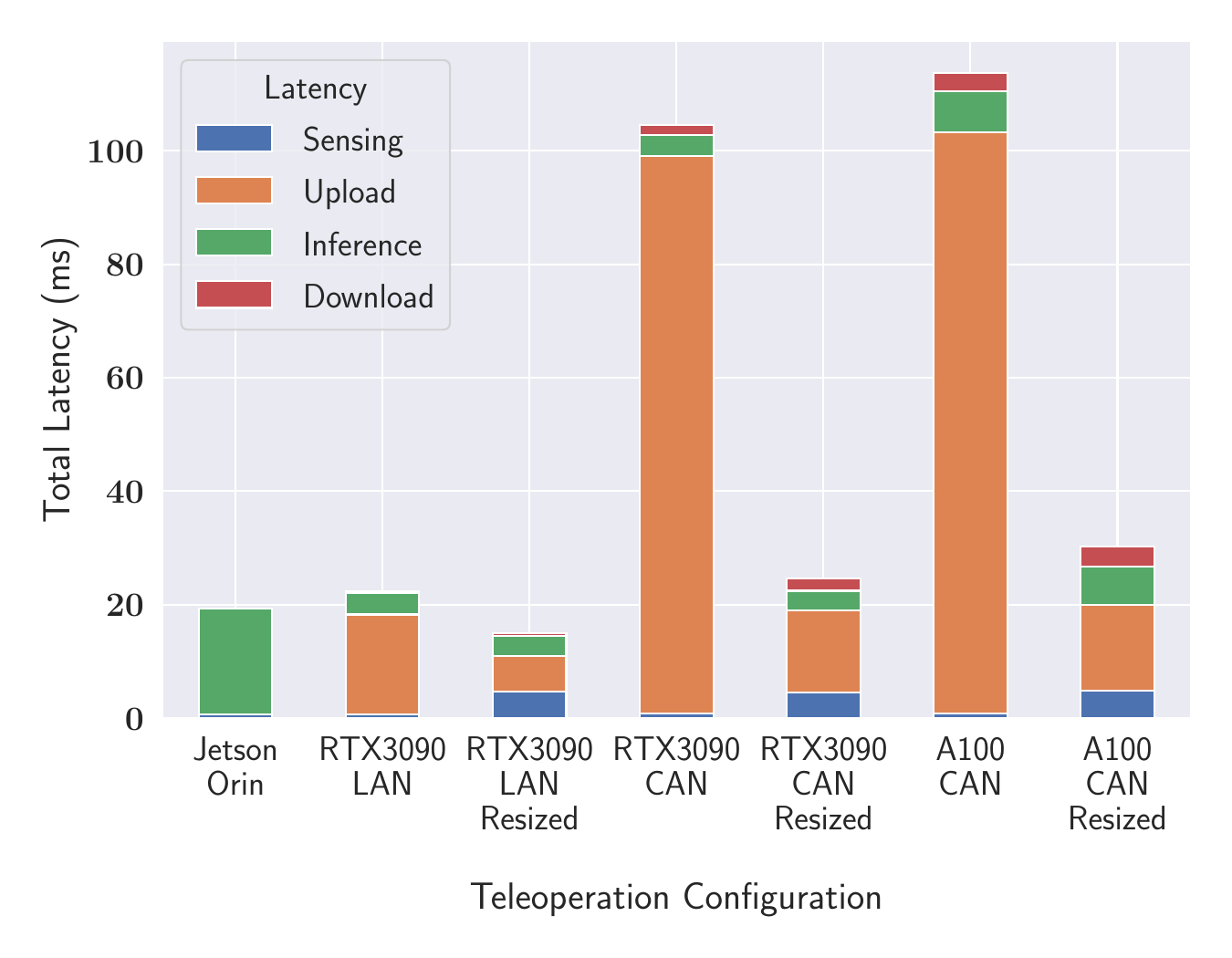}
    \caption{\small \textbf{Profiling with \name reveals latency tradeoffs and costs in end-to-end teleoperation pipelines.} \name identifies the combination of an RTX3090 GPU, local image resizing, and a LAN to be the most performative teleoperation configuration, with lower end-to-end latency than local inference.}
    \label{fig:franka-stacked-barplot}
    \vspace{-1em}
\end{figure}
\begin{table*}[t!]
\centering
\begin{tabular}
{| c | c | c | c | c | c | c | c | c |}
\hline
\multicolumn{3}{|c|}{\textbf{Configuration}} & \multicolumn{5}{c|}{\textbf{Latency (mean $\pm$ std in ms)}} \\ 
\hline
 \textbf{Resize Images} & \textbf{GPU}  & \textbf{Network} & \textbf{Sensing} & \textbf{Upload} & \textbf{Inference} & \textbf{Download} & \textbf{Total}\\
\hline
 \xmark & Jetson Orin & \xmark  & $0.71 \pm 0.26$ & \xmark              & $18.55\pm 3.34$   & \xmark                & $19.26 \pm 3.34$\\
 \xmark & RTX 3090 & LAN        & $0.76 \pm 0.25$ & $17.53 \pm 1.72$    & $3.73 \pm 1.12$   & $0.37 \pm 0.12$       & $22.38 \pm 2.07$\\
 \cmark & RTX 3090 & LAN        & $4.71 \pm 1.08$ & $6.25 \pm 0.99$     & $3.48 \pm 1.11$   & $0.62 \pm 0.11$       & $15.07 \pm 1.84$\\
 \xmark & RTX 3090 & CAN        & $0.81 \pm 0.28$ & $98.28 \pm 11.33$   & $3.73 \pm 1.53$   & $1.73 \pm 1.81$       & $104.55\pm 11.58$\\
 \cmark & RTX 3090 & CAN        & $4.50 \pm 1.03$ & $14.53 \pm 10.13$   & $3.43 \pm 1.52$   & $2.24 \pm 6.05$       & $24.71 \pm 11.94$\\
 \xmark & A100 & CAN            & $0.79 \pm 0.27$ & $102.48 \pm 26.40$  & $7.19 \pm 3.38$   & $3.16 \pm 8.95$       & $113.62\pm 28.08$\\
 \cmark & A100 & CAN            & $4.83 \pm 1.08$ & $15.13 \pm 7.63$    & $6.71 \pm 3.05$   & $3.59 \pm 4.83$       & $30.26 \pm 9.59$\\
 \cmark & RTX 3090 & AT\&T 5G   & $3.03 \pm 0.58$ & $432.71 \pm 34.14$  & $3.05 \pm 1.35$   & $228.84 \pm 149.83$   & $667.63\pm 153.67$\\
 \cmark & A100 & AT\&T 5G       & $3.14 \pm 0.48$ & $339.20 \pm 234.52$ & $4.56 \pm 15.68$  & $128.04 \pm 108.65$   & $474.94\pm 258.94$\\
 \hline
\end{tabular}
\caption{\small \textbf{\name's profiling of teleoperation configurations for a robotic manipulation task.} \name quantifies the tradeoffs between network latency and inference cost, disambiguates upload and download latency, and identifies the most performative setups.
}
\vspace{-2em}
\label{table:results}
\end{table*}

Figure \ref{fig:franka-stacked-barplot} shows component and end-to-end latency for LAN and CAN configurations. \name reveals several tradeoffs at play. First, a tradeoff between inference latency and network latency--remote devices show lower inference latency than the edge device, but uploading images comes with additional latency due to networks. Second, a tradeoff between network latency and sampling latency--when resizing images at the edge, network latency is reduced due to smaller payload size, but, sampling time increases due to the latency of image resizing. We also note an asymmetry in network delay across all teleoperation configurations. As images are uploaded but actions are downloaded, upload latency contributes significantly to end-to-end latency, while download latency is negligible. Based on \name's profiling, we see that the configuration that minimizes end-to-end latency is to perform inference on an RTX3090 GPU connected via LAN, and to resize images before transmission.
Table \ref{table:results} shows the details of all the tested configurations, including those with an AT\&T mobile 5G network. We note that the AT\&T 5G configuration incurs extreme network latency, with end-to-end latency up to 4 times worse than those on a CAN.

In summary, we design, implement, and profile three teleoperation configurations. We show that \name collects specific data on important steps in a teleoperation pipeline and that profiling using \name can be used to identify optimal choices for configuring cloud robotic setups.

\section{Discussion and Conclusion}
We present \name, an end-to-end and real-time benchmarking suite for networked robotics. Our package is easy to use, interfaces with industry-standard hardware and software, and pays special attention to profiling one-way delays in wireless networks. We demonstrate the modularity and scope of \name's profiling abilities through the exploration of two robotic offloading experiments and a teleoperation experiment on a Franka Emika Panda arm. We believe \name is a significant step towards the efficient deployment of optimized networked robotics.

\section{Acknowledgement}
This work was supported in part by the Lockheed Martin Corporation, in part by the National Science Foundation under grant 2148186, and in part by federal agencies and industry partners as specified in the Resilient and Intelligent NextG Systems (RINGS) Program. 

{
\bibliographystyle{ieeetr}
\bibliography{references}
}

\newpage
\appendix
\section{Appendix}
\subsection{Tabulated Metrics Collected by Offloading CLI}
\label{app:CLI}
Table \ref{tab:metrics} shows a list of metrics collected by the CLI for offloaded inference (Algorithm \ref{alg: offloaded-inference}) by default.

\begin{table}[ht!]
\centering
\begin{tabular}{l p{5cm}}
\toprule
\textbf{Metric} & \textbf{Description}\\ 
\midrule
Sensing Time & Time taken for a sensor $\mathcal{S}$ to return sample $x$\\
Upload Size & Size of serialized $x$ to be uploaded to $\mathcal{C}$\\
Upload Latency & Time taken for $\mathcal{D}$ to transmit $x$ to $\mathcal{S}$\\
Upload Throughput & Observed throughput on uploading $x$ to $\mathcal{S}$\\
Inference Latency & Time taken for $\mathcal{S}$ to perform inference with $\mathcal{M}$\\
Download Size & Size of serialized $y$ to be transmitted to $\mathcal{D}$\\
Download Latency & Time taken for $\mathcal{S}$ to transmit $y$ to $\mathcal{D}$\\
Download Throughput & Observed throughput on transmitting $y$ to $\mathcal{D}$\\
\bottomrule
\end{tabular}
\caption{CLI Options and Descriptions}
\label{tab:metrics}
\end{table}


\subsection{Terminal Commands for Offloaded Image Classification}
\label{app:terminal}
Below is an example of a terminal command for the offloading inference CLI. These commands reproduce the experiment with the small variant of EfficientNetV2, running between a server and a client. Table \ref{tab:cli_options} gives descriptions of the commands.

\lstset{basicstyle=\small\ttfamily}
\begin{lstlisting}[language=sh]
PEERNet --server --name cluster 
        --network zmq-tcp 
        --network-config net_config.yaml 
        --model-name efficientnet_v2_s 
        --device cuda:6 
        --iterations 10005
         
PEERNet --client --name jetson-orin 
        --network zmq-tcp 
        --network-config net_config.yaml 
        --dataset-loc sample-image.JPEG 
        --iterations 10005 
        --result-loc results 
        --generate-plots
\end{lstlisting}

\begin{table}[ht!]
\centering
\begin{tabular}{>{\ttfamily}l p{4.9cm}}
\toprule
\textbf{Option} & \textbf{Description} \\
\midrule
--server/--client       & Indicate device role.\\
--name             & Device name in network configuration. \\
--network               & [zmq-tcp $|$ zmq-udp $|$ ros] Specify the network type.\\
--network-config   & Path to the network configuration.\\
--sensor-type      &  Specify sensor.\\
--dataset-loc      & Location of the dataset.\\
--sensor-object    & Location of the sensor object.\\
--iterations    & Number of trials. \\
--result-loc  &  Output directory.\\
--model-name       & ML model name. \\
--device           &  cuda / cpu.\\
--generate-plots        &  Create output files.\\
\bottomrule
\end{tabular}
\caption{CLI Options and Descriptions}
\label{tab:cli_options}
\end{table}

\end{document}